\documentclass[twocolumn]{IEEEtran} 
\usepackage[utf8]{inputenc}

\title{Weakly-Supervised Semantic Segmentation of Ships Using Thermal Imagery}

\author{\IEEEauthorblockN{Rushil Joshi\IEEEauthorrefmark{1}\IEEEauthorrefmark{2},
Ethan Adams\IEEEauthorrefmark{1}\IEEEauthorrefmark{3}, Matthew Ziemann\IEEEauthorrefmark{4}\IEEEauthorrefmark{2} and
Christopher A. Metzler\IEEEauthorrefmark{2}}\\
\IEEEauthorrefmark{2} University of Maryland, College Park,
\IEEEauthorrefmark{3} University of North Texas,
\IEEEauthorrefmark{4} DEVCOM Army Research Laboratory\newline
\IEEEauthorrefmark{1} Equal Contributions\\
Email: rjoshi23@umd.edu, metzler@umd.edu}
\date{}

\usepackage{amsmath,amsfonts,amssymb}
\usepackage{mathtools}
\usepackage{xcolor}
\usepackage{mwe}
\usepackage{mathabx}
\usepackage{multirow}
\usepackage{bigstrut}
\usepackage{booktabs}
\usepackage{pgfgantt}
\usepackage{rotating}
\usepackage{tikz}
\usetikzlibrary{shapes,arrows}
\usetikzlibrary{snakes} 
\usepackage{pdflscape}
\usepackage{float}
\usepackage{listings}
\usepackage{lipsum}
\usepackage{graphicx}
\usepackage{wrapfig}
\usepackage{array,ragged2e}
\usepackage{multicol}
\usepackage{caption}
\usepackage{subcaption}
\usepackage{algorithm}
\usepackage[noend]{algpseudocode}
\usepackage{setspace}

\usepackage[colorlinks=true, allcolors=blue]{hyperref}
\usepackage{changepage}
\algnewcommand{\Inputs}[1]{%
	\State \textbf{Inputs:}
	\Statex \hspace*{\algorithmicindent}\parbox[t]{.8\linewidth}{\raggedright #1}
}
\algnewcommand{\Initialize}[1]{%
	\State \textbf{Initialize:}
	\Statex \hspace*{\algorithmicindent}\parbox[t]{.8\linewidth}{\raggedright #1}
}

\usepackage[style=numeric,sorting=none]{biblatex}
\usepackage[export]{adjustbox}

\begin{document}

\maketitle
\begin{abstract}

The United States coastline spans 95,471 miles \cite{shoreline}; a distance that cannot be effectively patrolled or secured by manual human effort alone. 
Unmanned Aerial Vehicles (UAVs) equipped with infrared cameras and deep-learning based algorithms represent a more efficient alternative for identifying and segmenting objects of interest---namely, ships. 
However, standard approaches to training these algorithms require large-scale datasets of densely labeled infrared maritime images. Such datasets are not publicly available and manually annotating every pixel in a large-scale dataset would have an extreme labor cost.

In this work we demonstrate that, in the context of segmenting ships in infrared imagery, weakly-supervising an algorithm with sparsely labeled data can drastically reduce data labeling costs with minimal impact on system performance. 
We apply weakly-supervised learning to an unlabeled dataset of 7055 infrared images sourced from the Naval Air Warfare Center Aircraft Division (NAWCAD). 
We find that by sparsely labeling only 32 points per image, weakly-supervised segmentation models can still effectively detect and segment ships, with a Jaccard score of up to 0.756. 

\end{abstract}


\section{Introduction}

Deep object detection models have demonstrated the ability to find and classify objects of interest - an essential asset to the interest of national security \cite{projectmaven}.
Due to the breadth of the United States coastline, ships and other watercraft are among the highest-priority objects of interest. Automating the task through the application of object detection networks allows for efficient, yet accurate detection at comparable-to-human accuracy. 


Neural networks represent the state-of-the-art solution for identifying objects of interest for UAV operators in the thermal domain, which is commonly used in maritime environments. Deep convolutional neural networks are able to detect objects of interest and segment image into individual classes with high accuracy in the visual spectrum, as well as the infrared spectrum \cite{ObjectDetectionThermal}\cite{birdsAI}. Image segmentation offers greater detail than detection within each prediction, since each pixel has an associated predicted class. In addition to visible spectrum images, neural-based segmentation has shown quality results for thermal images \cite{ImageSegThermal,ThermalSurvey} as well.



\begin{figure}[t]
\centering
  \begin{subfigure}[c]{.45\linewidth}
  \centering
    \includegraphics[width=\linewidth]{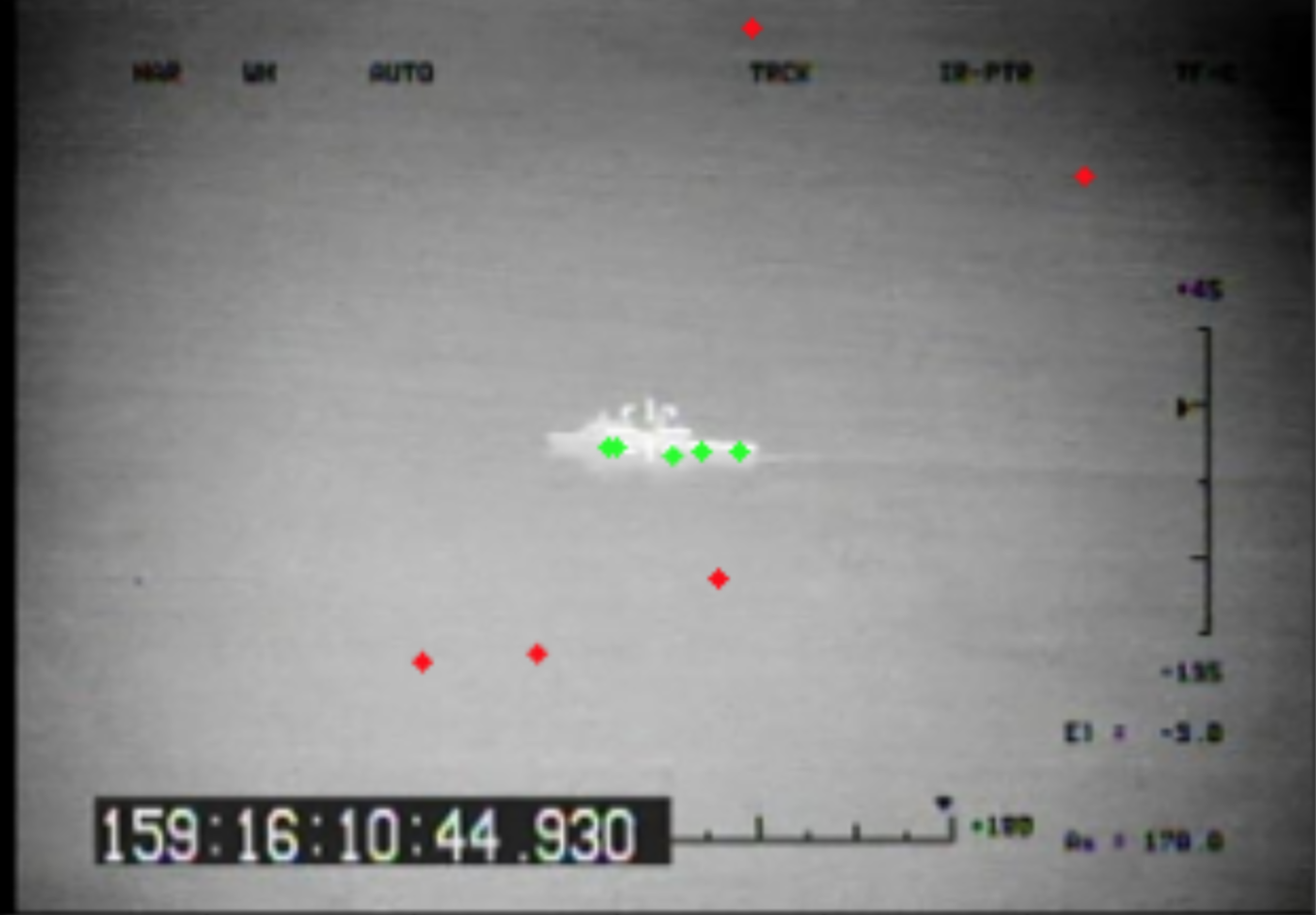}%
    \end{subfigure}
    ~
  \begin{subfigure}[c]{.45\linewidth}
  \centering
    \includegraphics[width=\linewidth]{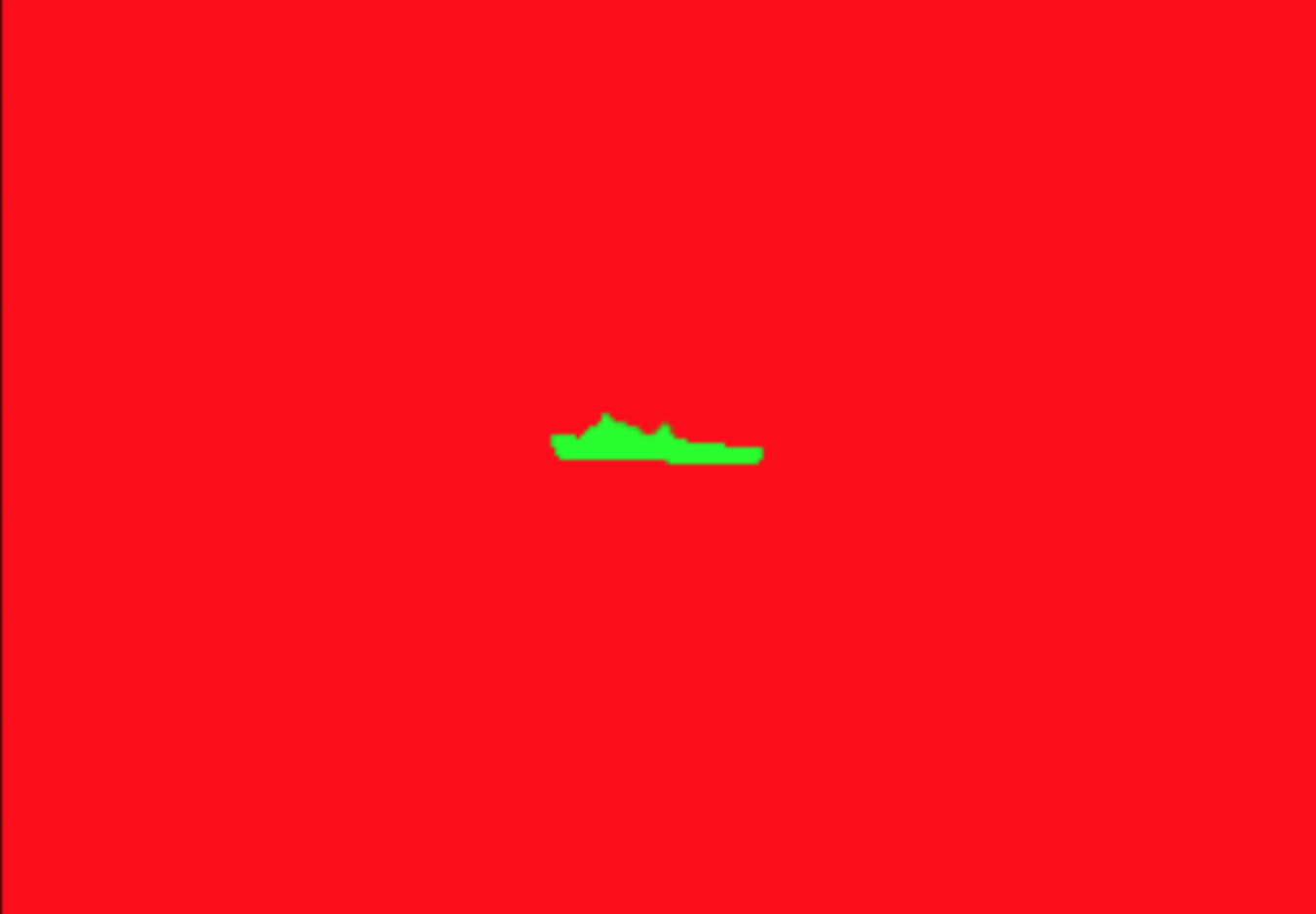}%
    \end{subfigure}
  \caption
    {%
    \small
    Sparse annotations (left) versus dense annotations (right). Sparse annotation are far easier to generate.
      \label{fig:pointvseg}%
    }
\end{figure}

Traditionally, segmentation methods require ground truth annotations in the form of masks, marking each pixel in the image with an object classification. However, this approach requires an extensive amount of time and human resources. An example of a full segmentation mask is displayed in Figure \ref{fig:pointvseg}. 
The introduction of weakly-supervised learning \cite{Pointly,whatsthepoint,pointsupercrowd,cropspoint}, allows for annotators to quickly and efficiently perform sparse annotation rather than annotating fully-segmented mask, without compromising on segmentation performance. The success of image segmentation trained on weakly-supervised annotations dispels standard of requiring fully-segmented mask annotations. 



This paper expands on the usefulness of the proposed methods --- namely, point-wise and squiggle-wise annotation --- to annotate and train a model for segmentation using weakly-labeled infrared images. By utilizing a modified U-Net segmentation architecture and point supervision on thermal images, we illustrate the effectivity of using point-wise annotations to utilize object segmentation with minimal truth labels, and therefore minimal annotation work. We also demonstrate that this method is robust to constraints presented in our study, including but not limited to poor image quality, noise due to varying environmental conditions, a minimal number of annotations, and limited training times.

\section{Related Work}

\subsection{Infrared Domain Image Segmentation}
Image segmentation is among the most popular and widely-researched computer vision problems, across a wide variety of imaging domains, including the visible spectrum (RGB) and infrared (IR) domains.  State-of-the-art neural-based image segmentation solutions generally make use of an encoder-decoder architecture, often with additional feature fusion modifications such as skip connections and residuals. In the encoder, input images are reduced to a low-resolution feature space, with generalized features learned earlier and image-specific features learned further into the network. The decoder learns to up-sample latent features learned by the encoder and project them back into a segmentation mask, for which each pixel is assigned a class probability. To preserve spatial information within the network, architectures such as U-Net \cite{unet} and SegNet \cite{segnet} utilize skip connections, concatenating previously-learned features to latter layer inputs. Other state-of-the-art segmentation networks, such as Mask R-CNN \cite{maskrcnn}, use a pre-trained backbone convolutional neural network in tandem with a region proposal network to learn latent feature mappings, which are then used to generate a segmentation mask.

Specifically for images in the IR domain, skip-enabled encoder-decoder architectures have a history of success for various segmentation tasks. \cite{photovoltaicSeg} uses a modified U-Net architecture with residual learning capability for the task of segmenting photovoltaic panels within thermal images, scoring a Jaccard index score of 92.65\% with a standard U-Net and over 94\% using the modified architecture.  Furthermore, \cite{ImageSegThermal} provides a comprehensive comparative analysis across various thermal segmentation datasets, along with performance metrics from U-Net and other similar segmentation architectures (SegNet, MFNet). 

While state-of-the-art approaches generally perform well on most image segmentation datasets, even with a limited number of samples, annotation of segmentation masks is significantly more costly when considering the time required to annotate a single image. Large, generalizable segmentation datasets that are fully and accurately painted with labels, such as MS-COCO \cite{MSCOCO}, demonstrably allow a network to learn large-scale features that can be used to pre-train for a smaller, more specific task, as shown in \cite{Pointly}. However, annotating for a custom segmentation task can take exponentially longer amounts of time as the number of classes and complexity of the images increases.
\subsection{Weakly-Supervised Segmentation}
Due to the significant increase in cost of full segmentation mask annotation compared to other vision-based tasks, existing studies have sought to mitigate the time required to annotate while simultaneously not compromising performance. Bearman et al. introduce the concept of image-level, squiggle-level, and point-level supervised segmentation, along with the average annotation time and evaluation performance for each type \cite{whatsthepoint}. The study performed three-types of weakly-supervised annotation schemes on the Pascal-VOC dataset, and compared their annotation times and performance to full segmentation masks. Point-level supervision requires one or more points annotated per class object, as opposed to full pixel-accurate painting, taking an average of 22.1 seconds per image. Squiggle-level supervision, similarly, requires one or more “squiggles”, or continuous series of points per image, averaging 34.9 seconds per image. In comparison, full segmentation mask painting took an average of 239.7 seconds per image - nearly 7 times slower than squiggle-level annotation and over 10 times slower than point-level annotation.
\subsection{Point-Supervised Segmentation}
One method of weakly-supervised segmentation only requires the annotation of only a small, variable number of points to be annotated, rather than painted over entirely. Point-supervised segmentation, namely, has been supported by recent research to be a viable solution to the costly process of segmentation annotation. In addition to the study performed by Bearman et al. \cite{whatsthepoint}, Cheng et al. outline a simple, yet effective point annotation scheme, where only 10 points are selected and classified within a bounding box of an object \cite{Pointly}. Trained on Mask R-CNN and evaluated only on annotated points, this annotation scheme achieved up to 98\% of its fully-supervised performance, despite being nearly 5 times faster to annotate point-wise. Cheng et al. further show that applying the paradigm of transfer learning by pretraining on existing large-scale segmentation data prior to training on point annotations is equally effective - both in the case of full masks of the larger dataset being available, as well as when only point annotations of said dataset are available \cite{Pointly}.
\section{Method}

\subsection{Datasets}\label{section:dataset}
\begin{figure*}[ht]
    \centering
    	\begin{subfigure}[]{0.2\textwidth}
    		\includegraphics[width=\linewidth]{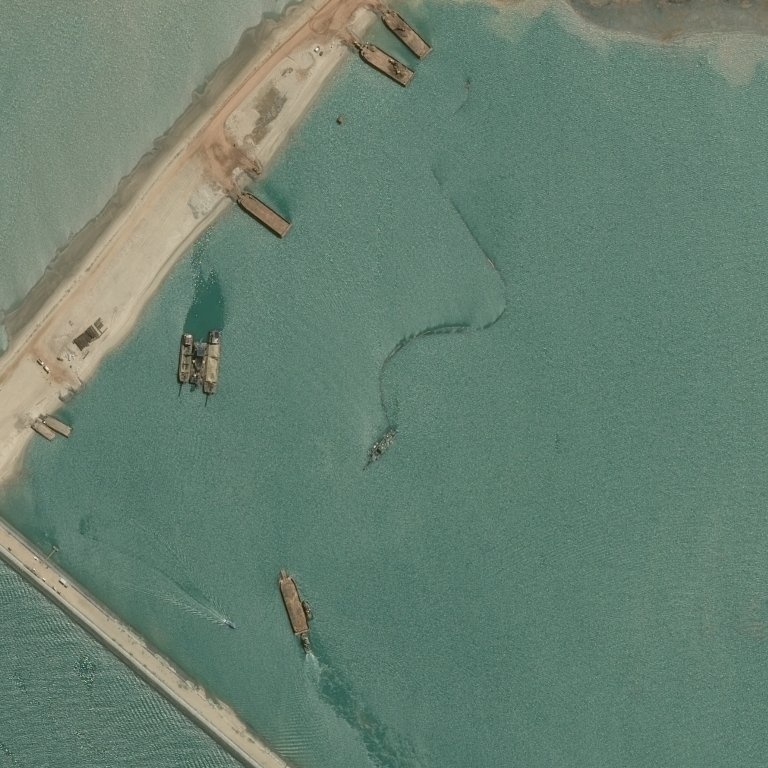}
    	\end{subfigure}\hfill
    	\begin{subfigure}[]{0.2\textwidth}
    		\includegraphics[width=\linewidth]{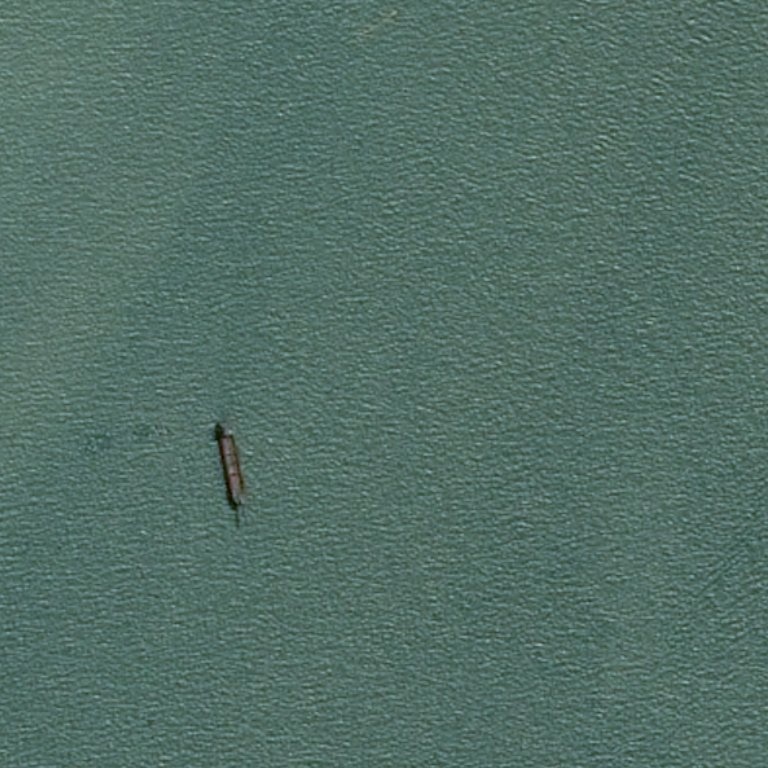}
    	\end{subfigure}\hfill
    	\begin{subfigure}[]{0.2\textwidth}
     		\includegraphics[width=\linewidth]{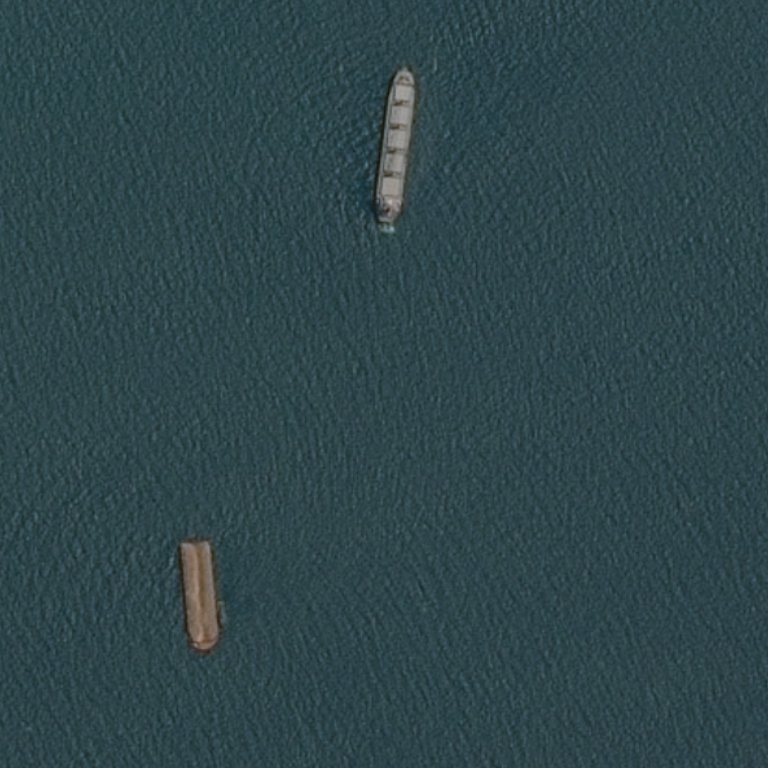}
    	\end{subfigure}\hfill
    	\begin{subfigure}[]{0.2\textwidth}
     		\includegraphics[width=\linewidth]{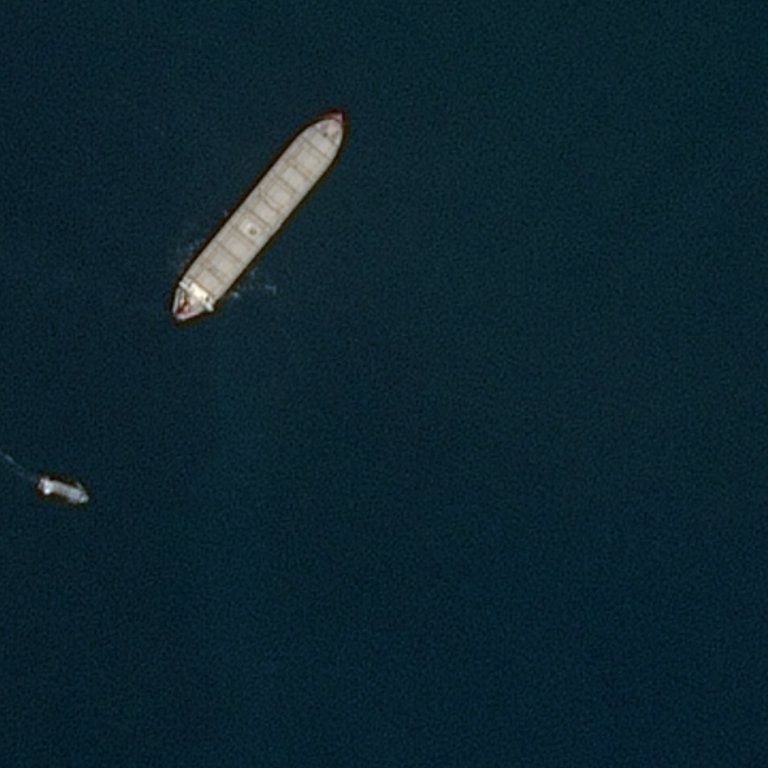}
    	\end{subfigure}\\
    		\caption{\small Example images from the Airbus dataset~\cite{AIRBUS}. These satellite images are clean of any post-processing markings, ships are visually identifiable, and the intereference from structures or landmasses is low.}
    		\label{fig:AIRBUSSAMPLES}
\vspace{25pt}
    \begin{subfigure}[]{0.2\textwidth}
        \includegraphics[width=\linewidth]{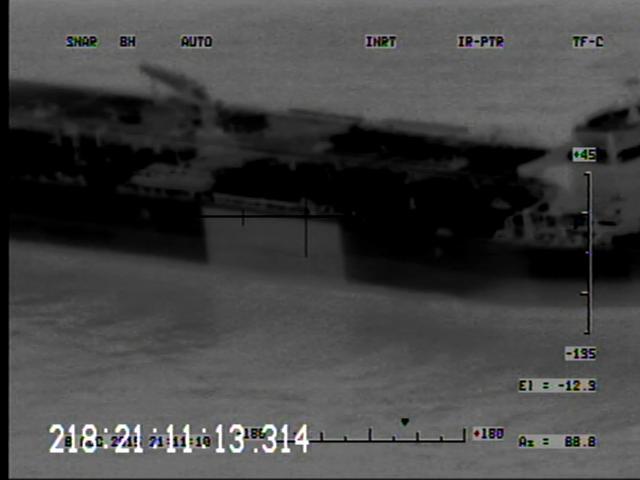}
    \end{subfigure}\hfill
    \begin{subfigure}[]{0.2\textwidth}
        \includegraphics[width=\linewidth]{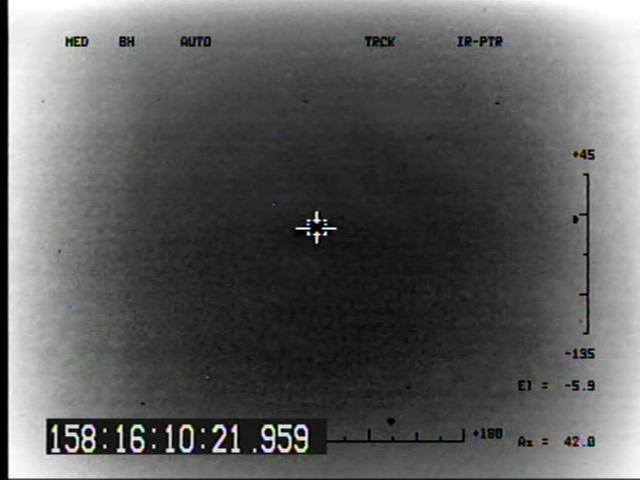}
    \end{subfigure}\hfill
    \begin{subfigure}[]{0.2\textwidth}
        \includegraphics[width=\linewidth]{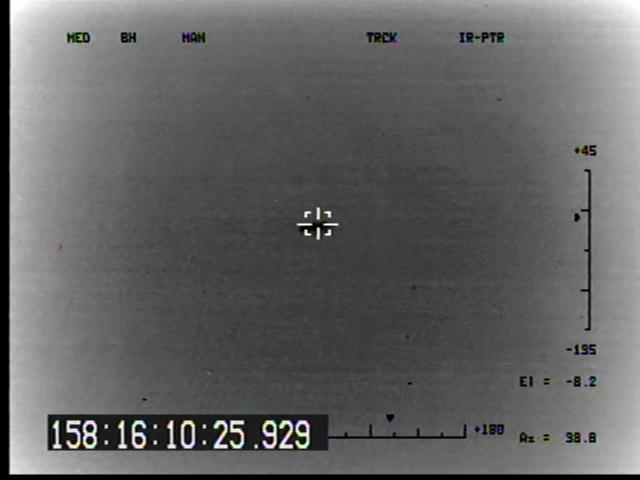}
    \end{subfigure}\hfill
    \begin{subfigure}[]{0.2\textwidth}
        \includegraphics[width=\linewidth]{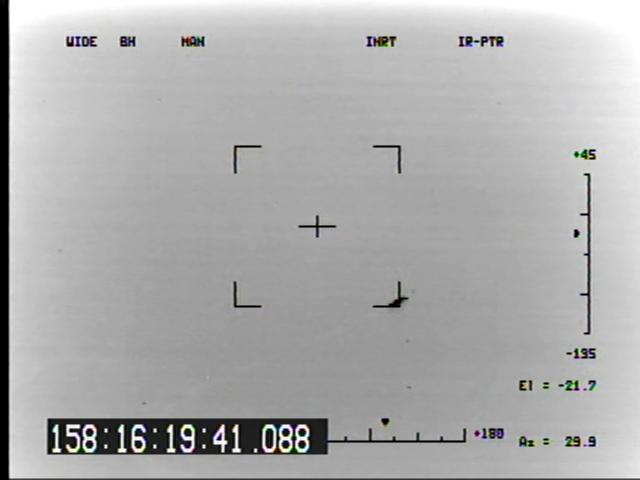}
    \end{subfigure}\hfill
    \caption{\small Example images from the NAWCAD dataset. These images contain post processing marks such as scales, crosshairs, and time stamps imposed onto the image.}.
    \label{fig:NAWCADSAMPLES}
\end{figure*}

\begin{figure}[ht]
    \centering
    \begin{subfigure}[]{0.2\textwidth}
    \centering
        \includegraphics[width=.95\linewidth]{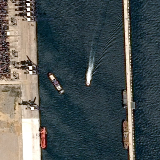}
        \caption{Input Image}
    \end{subfigure}
    \begin{subfigure}[]{0.2\textwidth}
    \centering
        \includegraphics[width=.95\linewidth]{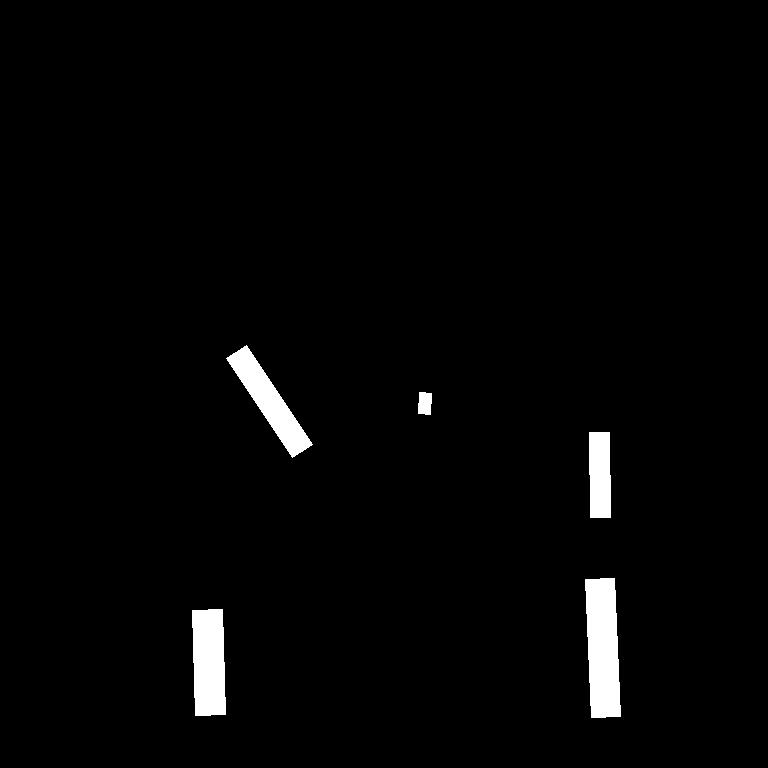}
        \caption{No Mask}
    \end{subfigure}\vspace{10pt}
    \begin{subfigure}[]{0.2\textwidth}
    \centering
        \includegraphics[width=.95\linewidth]{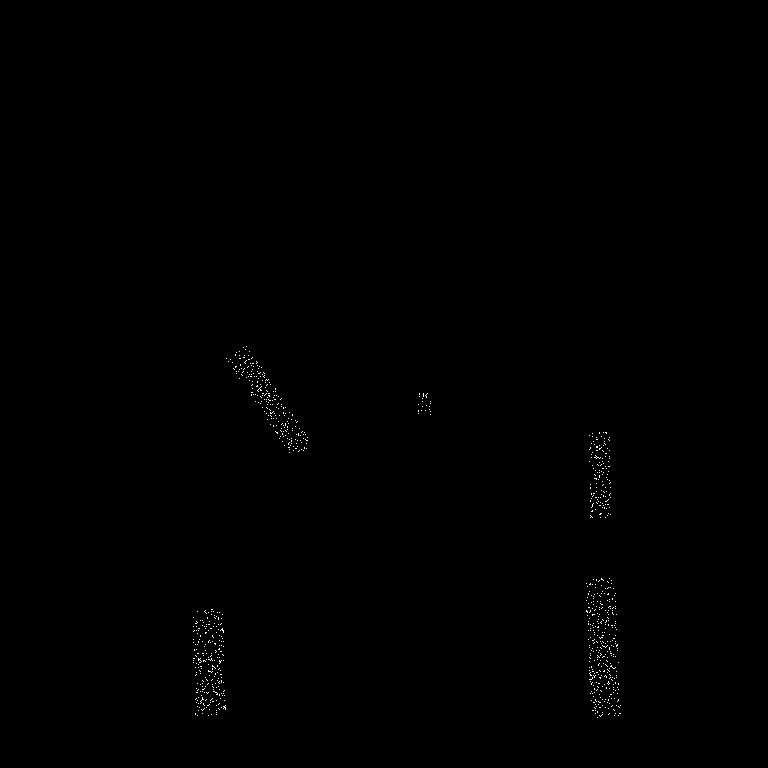}
        \caption{90\% Mask}
    \end{subfigure}
    \begin{subfigure}[]{0.2\textwidth}
    \centering
        \includegraphics[width=.95\linewidth,height=.95\linewidth]{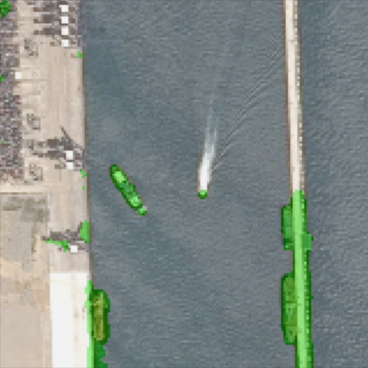}
        \caption{Prediction}
    \end{subfigure}
    
	\caption{\small We show the different configurations of training labels that were used for testing the hypothesis of using subsampled data to train a segmentation network.}    \label{fig:AirBusReduced}
\end{figure}

     We use two primary datasets, depicted in Table \ref{tab:dataset breakdown}, to explore weakly-supervised learning and its application to ship segmentation:
        \begin{table}[ht]
        \centering
        \begin{tabular}{c|c|c|c|c}
             Dataset&Domain&\# Images& Resolution&Label Scheme\\
             \hline
             Airbus&Visible&~200,000&780x780&Full mask  \\
             NAWCAD&Infrared&~7,000&640x480&Unlabeled 
        \end{tabular}
        \caption{\small Details of the two datasets used during training.}
        \label{tab:dataset breakdown}
    \end{table}\\
     The Airbus dataset \cite{airbus}, consists of clear, high-resolution RGB images taken from a flat aerial perspective, shown in Figure \ref{fig:AIRBUSSAMPLES}. The NAWCAD (Naval Air Warfare Center Aircraft Division) dataset, however, includes comparatively-lower resolution infrared images, with only a fraction of the volume of the Airbus set. The NAWCAD dataset contains 7055 unlabeled infrared images taken from an unmanned aerial vehicle, in both white-hot and black-hot formats. In contrast to the Airbus images, NAWCAD images were captured with varying angles and environmental conditions. All NAWCAD images were collected from the bay area of the Patuxent River in Maryland. A sample of the dataset is displayed in Figure \ref{fig:NAWCADSAMPLES}. 
    Given the generally-limited availability of infrared images, along with the additional limitations of the NAWCAD images, we aimed to utilize the external Airbus dataset to supplement the weakly-supervised learning process, via transfer learning. 
    In order to determine a baseline potential for weakly-supervised segmentation, we train with a random sample of mask pixels from the fully-segmented Airbus set.  Each pixel is labeled as either boat or background. The dataset contains some overlap between nearby boats, landmasses, or other objects, but the majority of objects are isolated in open water.
    We randomly mask \(n\%\) of all pixel labels, at both the  both at n=95\% and n=90\% levels.
    The generated masks were then sent through the data pipeline as the ground truth for the model.
    A comparison of full labeled data (no mask) to a 95\% mask can be seen in Figure \ref{fig:AirBusReduced}.

\subsection{Data Preparation}
\begin{figure}[t]
        \centering
        \begin{subfigure}{.2\textwidth}
            \includegraphics[width=\textwidth]{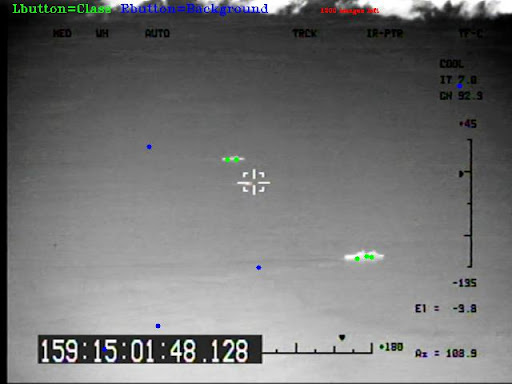}
            \caption{N=10 Point Click}

        \end{subfigure}\hfill
        \begin{subfigure}
            {.2\textwidth}
            \includegraphics[width=\textwidth]{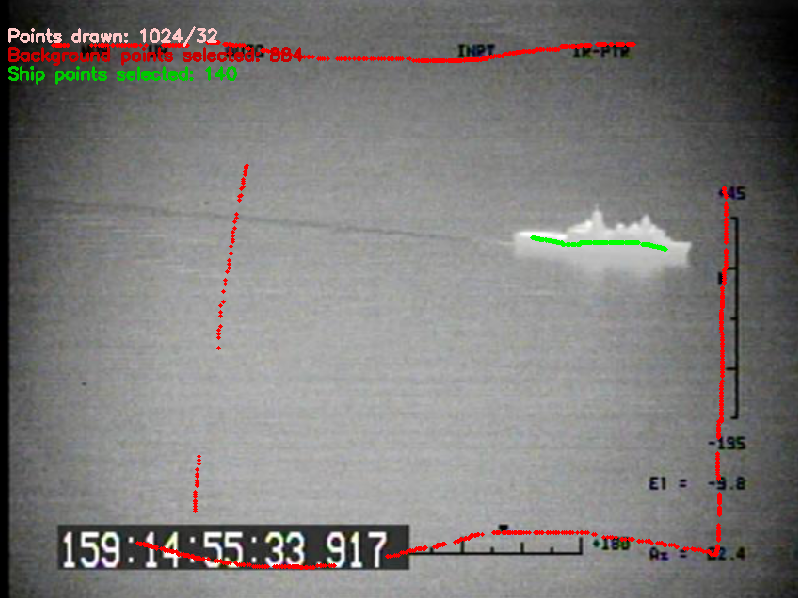}
            \caption{N=32 Squiggle}
        \end{subfigure}
        \caption{\small Example displaying the different annotation schemes used to create labels. (a) Shows the point annotation method where 10 points are sampled with an even distribution between ship and background. (b) Shows the use of drawing lines/squiggles on the image, from which we extract 32 random points weighted by the proportion of points drawn per class.}
        \label{fig:Selection_Vis}
    \end{figure}

    Inspired by the results using weak supervision on the Airbus dataset, we explore two annotation schemes for labeling the NAWCAD IR dataset, shown in Figure \ref{fig:Selection_Vis}. The first method - \textit{point} annotation - involves selecting 5 points for each class (namely, the ship and background class) to allow for an even sample distribution across all classes. The second method - \textit{squiggle} annotation - involves drawing squiggle lines on each class, from which a weighted random sample of 32 points are selected. Squiggle annotation allows for a larger, more scaled sample of points to be considered while training, while still taking around the same amount of time as point annotation. We annotated 1200 infrared images using the point-supervised and squiggle-supervised annotation schemes. 
    Once all images were annotated using the respective methods, the new dataset was randomly shuffled and then split into 90\% training / 10\% validation sets, due to limited amount of trainable images.
    Prior to training, the dataset is augmented via standard segmentation transformations, including random shuffling, rotation, scaling, and cropping.
    \subsection{Loss Function}
     
\begin{figure}[t]
    \centering
    \includegraphics[width=.5\textwidth]{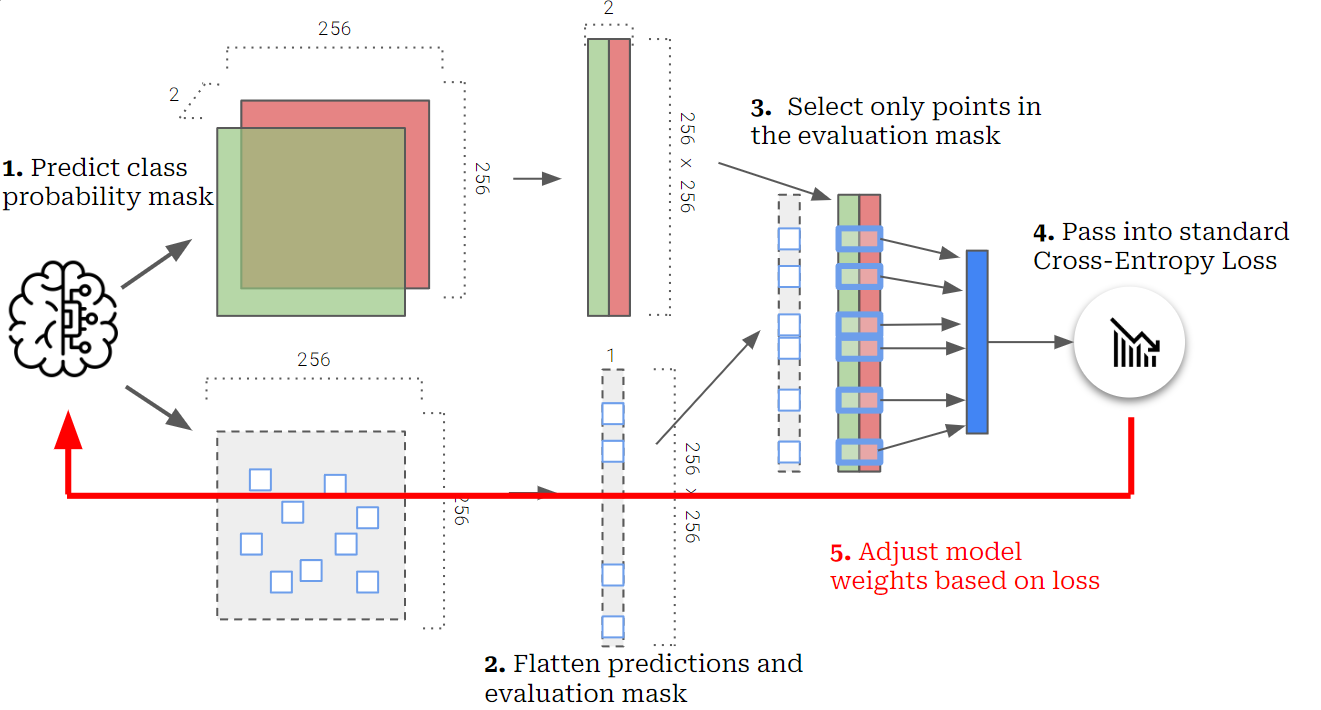}
    \caption{\small This diagram visually illustrates how the labels and evaluation mask are applied to the loss function during training.}
    \label{fig:PWCE}
\end{figure}

    Pixel-Wise Cross-Entropy (PWCE) was used to train both annotation schemes. PWCE is an implementation inspired by the original study introducing point-wise supervision \cite{Pointly}, with our method of point sampling in mind. 
    \begin{equation}
        L_{PWCE}(Y, \hat{Y}, m) = -\sum_{i=1}^{n} [(Y_im_i)*log(\hat{Y_i}m_i)]
    \end{equation}
    A diagram of the process is shown in figure \ref{fig:PWCE}.
    
    During the generation of the dataset, the loader creates three different images; the original image, the ground truth annotated points, and an evaluation mask denoting which points in the ground truth label to evaluate on (represented as a sparse boolean array). This way, points not to be considered in evaluation are zeroed out by the mask, and not factored into the loss function. 
    		

\subsection{Transfer Learning}
    Due to the limitations of the NAWCAD dataset, as mentioned in \ref{section:dataset}, we explore the effectiveness of transfer learning in combination with augmentation techniques. This would ideally allow the network to better generalize previous strongly-learned features of interest to our weakly-supervised task, along with reducing total training time. We applied transfer learning for both point-supervised and squiggle-supervised annotation schemes. The model was first pretrained on the full Airbus dataset, with grayscaling and random inversion augmentations applied to account for the domain change to both formats of thermal images. The goal of doing this was to assist the model with identifying boat shapes and edges in contrast with the background scenerary. 

\section{Results}
This section demonstrates the difference in performance when utilizing 10 evenly distributed samples versus 32 randomly distributed samples. For our anaylsis we created a subset of fully segmented images that were not used during training for inference. The predictions are scored against the inference labels by measuring the precision, recall and Jaccard index score. The Jaccard score gives us quantitave metric of accuracy that measures the ratio of correct pixel-wise classifications to the total number of classifications. It is defined as:

\[ J(A, B) = \frac{|A\cup B|}{|A|+|B|-|A \cap B|}\]

\subsection{32-point Subsample Results}
The success cases from the squiggle annotations of the IR dataset are displayed in the last row of Table \ref{tab:images}. An outline of the training performance is shown in Tables \ref{tab:pretrained} and \ref{tab:not-pretrained}. We observe that even with a massive reduction in trainable labels, the performance of the model is still admirable. However, as displayed in the last two columns of Table \ref{tab:images}, due to the low amount of overall training image variety, eliminating false positives completely and achieving full segmentation from the model still pose a problem, especially in heavy atmospheric conditions.  

\subsection{10-point Subsample Results}
The analysis for this training scheme is depicted in Tables \ref{tab:pretrained} and \ref{tab:not-pretrained}. These tables depict that with further reduction of the trainable labels, the model still performs relatively similar to the 32-point training regime. The second row of Table \ref{tab:images} displays a visual representation of predictions generated during inference. The first three columns show the success cases which our model perfomed optimally; these images show that with little to no environmental noise, a model trained on only pointly-annotated labels performs well. However, as seen in the last two columns, the model begins to falter and over segment in the presence of boat wake trails and cloud coverage. In addition, we observe that segmentation tends to falter on large ships that extend beyond the frame of the image.
\section{Conclusion}
In this work, we illustrate that a weakly-supervised annotation scheme can be effectively used to perform thermal image segmentation in high-noise, maritime environments, from only several hundred training samples. Despite taking substantially less time for annotation compared to full segmentations, point-supervised segmentation can be used to successfully and robustly detect and segment both large and small ships. We also investigate the effect of transfer learning from full segmentations in the visible domain (RGB) on weakly-supervised segmentation, finding that pre-training on similar ship image data marginally improves performance after training for the same number of epochs. While we still observed certain phenomena (such as clouds or boat trails) affecting detection performance on occasion, this was not the case the majority of the time. As a part of future work, annotating and training on more images containing prevalent cloud coverage and boat trails would likely improve detection accuracy. \footnote{Link to code and data: \href{https://github.com/rushiljoshi/Weakly-Supervised-Ship-Seg}{Github}}
\newcommand{\scales}{.17}

\section*{Acknowledgements}
This work was supported by the University of Maryland Applied Research Lab for Intelligence and Security (ARLIS) and the Naval Air Warfare Center Aircraft Division (NAWCAD), as a part of the Research for Intelligence and Security Challenges (RISC) internship program. Metzler was supported in part by the AFOSR Young Investigator Program.
\nocite{*}
\printbibliography

@article{Pointly,
  author    = {Bowen Cheng and
               Omkar Parkhi and
               Alexander Kirillov},
  title     = {Pointly-Supervised Instance Segmentation},
  journal   = {CoRR},
  volume    = {abs/2104.06404},
  year      = {2021},
  url       = {https://arxiv.org/abs/2104.06404},
  eprinttype = {arXiv},
  eprint    = {2104.06404},
  bibsource = {dblp computer science bibliography, https://dblp.org}
}

@article{DBLP:journals/corr/JegouDVRB16,
  author    = {Simon J{\'{e}}gou and
               Michal Drozdzal and
               David V{\'{a}}zquez and
               Adriana Romero and
               Yoshua Bengio},
  title     = {The One Hundred Layers Tiramisu: Fully Convolutional DenseNets for
               Semantic Segmentation},
  journal   = {CoRR},
  volume    = {abs/1611.09326},
  year      = {2016},
  url       = {http://arxiv.org/abs/1611.09326},
  eprinttype = {arXiv},
  eprint    = {1611.09326},
  bibsource = {dblp computer science bibliography, https://dblp.org}
}

@misc{shoreline, title={How long is the U.S. shoreline?}, url={https://oceanservice.noaa.gov/facts/shorelength.html#:~:text=NOAA's\%20official\%20value\%20for\%20the,U.S.\%20shoreline\%20is\%2095\%2C471\%20miles}, journal={NOAA's National Ocean Service}, publisher={US Department of Commerce}, year={2010}, month={Jul}}

@misc{maskrcnn,
  doi = {10.48550/ARXIV.1703.06870},
  
  url = {https://arxiv.org/abs/1703.06870},
  
  author = {He, Kaiming and Gkioxari, Georgia and Dollár, Piotr and Girshick, Ross},
  
  title = {Mask R-CNN},
  
  publisher = {arXiv},
  
  year = {2017},
  
  copyright = {arXiv.org perpetual, non-exclusive license}
}

@article{unet,
  author    = {Olaf Ronneberger and
               Philipp Fischer and
               Thomas Brox},
  title     = {U-Net: Convolutional Networks for Biomedical Image Segmentation},
  journal   = {CoRR},
  volume    = {abs/1505.04597},
  year      = {2015},
  url       = {http://arxiv.org/abs/1505.04597},
  eprinttype = {arXiv},
  eprint    = {1505.04597},
  bibsource = {dblp computer science bibliography, https://dblp.org}
}

@misc{segnet,
  doi = {10.48550/ARXIV.1511.00561},
  
  url = {https://arxiv.org/abs/1511.00561},
  
  author = {Badrinarayanan, Vijay and Kendall, Alex and Cipolla, Roberto},
  
  title = {SegNet: A Deep Convolutional Encoder-Decoder Architecture for Image Segmentation},
  
  publisher = {arXiv},
  
  year = {2015},
  
  copyright = {arXiv.org perpetual, non-exclusive license}
}

@InProceedings{whatsthepoint,
author="Bearman, Amy
and Russakovsky, Olga
and Ferrari, Vittorio
and Fei-Fei, Li",
editor="Leibe, Bastian
and Matas, Jiri
and Sebe, Nicu
and Welling, Max",
title="What's the Point: Semantic Segmentation with Point Supervision",
booktitle="Computer Vision -- ECCV 2016",
year="2016",
publisher="Springer International Publishing",
address="Cham",
pages="549--565",
isbn="978-3-319-46478-7"
}

@inproceedings{NEURIPS2019_993edc98,
 author = {Sauder, Jonathan and Sievers, Bjarne},
 booktitle = {Advances in Neural Information Processing Systems},
 editor = {H. Wallach and H. Larochelle and A. Beygelzimer and F. d\textquotesingle Alch\'{e}-Buc and E. Fox and R. Garnett},
 pages = {},
 publisher = {Curran Associates, Inc.},
 title = {Self-Supervised Deep Learning on Point Clouds by Reconstructing Space},
 url = {https://proceedings.neurips.cc/paper/2019/file/993edc98ca87f7e08494eec37fa836f7-Paper.pdf},
 volume = {32},
 year = {2019}
}

@article{pointsupercrowd,
title = {Scale-Recursive Network with point supervision for crowd scene analysis},
journal = {Neurocomputing},
volume = {384},
pages = {314-324},
year = {2020},
issn = {0925-2312},
doi = {https://doi.org/10.1016/j.neucom.2019.12.070},
url = {https://www.sciencedirect.com/science/article/pii/S0925231219317795},
author = {Zihao Dong and Ruixun Zhang and Xiuli Shao and Yumeng Li}
}

@article{cropspoint, title={Active learning with point supervision for cost-effective panicle detection in cereal crops}, volume={16}, DOI={10.1186/s13007-020-00575-8}, number={1}, journal={Plant Methods}, author={Chandra, Akshay L. and Desai, Sai Vikas and Balasubramanian, Vineeth N. and Ninomiya, Seishi and Guo, Wei}, year={2020}}

@misc{WeakSupervisionPara,
  author       = "Alex, Ratner and Paroma, Varma",
  title        = "Weak supervision: A new programming paradigm for Machine Learning",
  howpublished = "http://ai.stanford.edu/blog/weak-supervision/",
  month        = "Mar",
  year         = "2019",
  journal      = "SAIL BLOG",
}

@article{DBLP:journals/corr/abs-1901-08396,
  author    = {Jonathan Sauder and
              Bjarne Sievers},
  title     = {Context Prediction for Unsupervised Deep Learning on Point Clouds},
  journal   = {CoRR},
  volume    = {abs/1901.08396},
  year      = {2019},
  url       = {http://arxiv.org/abs/1901.08396},
  eprinttype = {arXiv},
  eprint    = {1901.08396},
  bibsource = {dblp computer science bibliography, https://dblp.org}
}

@article{DBLP:journals/corr/abs-1801-08268,
  author    = {Utsav B. Gewali and
              Sildomar T. Monteiro},
  title     = {A Tutorial on Modeling and Inference in Undirected Graphical Models
              for Hyperspectral Image Analysis},
  journal   = {CoRR},
  volume    = {abs/1801.08268},
  year      = {2018},
  url       = {http://arxiv.org/abs/1801.08268},
  eprinttype = {arXiv},
  eprint    = {1801.08268},
  bibsource = {dblp computer science bibliography, https://dblp.org}
}

@misc{bigironsphere_2021, title={Loss function library - keras \&amp; pytorch}, url={https://www.kaggle.com/code/bigironsphere/loss-function-library-keras-pytorch/notebook}, journal={Kaggle}, publisher={Kaggle}, author={Bigironsphere}, year={2021}, month={Jul}}

@misc{airbus, title={Airbus Ship Detection Challenge}, url={https://www.kaggle.com/competitions/airbus-ship-detection/data}, journal={Kaggle}, publisher={Kaggle}, author={}, year={2018}, month={Jul}}

@inbook{photovoltaicSeg,
author = {Zhang, Hao and Hong, Xianggong and Zhou, Shifen and Wang, Qingcai},
year = {2019},
month = {10},
pages = {611-622},
title = {Infrared Image Segmentation for Photovoltaic Panels Based on Res-UNet},
isbn = {978-3-030-31653-2},
doi = {10.1007/978-3-030-31654-9_52}
}

@article{DBLP:journals/corr/abs-1807-03748,
  author    = {A{\"{a}}ron van den Oord and
              Yazhe Li and
              Oriol Vinyals},
  title     = {Representation Learning with Contrastive Predictive Coding},
  journal   = {CoRR},
  volume    = {abs/1807.03748},
  year      = {2018},
  url       = {http://arxiv.org/abs/1807.03748},
  eprinttype = {arXiv},
  eprint    = {1807.03748},
  bibsource = {dblp computer science bibliography, https://dblp.org}
}

@InProceedings{ObjLocalFree,
author = {Oquab, Maxime and Bottou, Leon and Laptev, Ivan and Sivic, Josef},
title = {Is Object Localization for Free? - Weakly-Supervised Learning With Convolutional Neural Networks},
booktitle = {Proceedings of the IEEE Conference on Computer Vision and Pattern Recognition (CVPR)},
month = {June},
year = {2015}
}

@article{PositiveUnlabeled,
  author    = {Yuewei Yang and
              Kevin J. Liang and
              Lawrence Carin},
  title     = {Object Detection as a Positive-Unlabeled Problem},
  journal   = {CoRR},
  volume    = {abs/2002.04672},
  year      = {2020},
  url       = {https://arxiv.org/abs/2002.04672},
  eprinttype = {arXiv},
  eprint    = {2002.04672},
  bibsource = {dblp computer science bibliography, https://dblp.org}
}

@article{DBLP:journals/corr/abs-2104-12763,
  author    = {Aishwarya Kamath and
              Mannat Singh and
              Yann LeCun and
              Ishan Misra and
              Gabriel Synnaeve and
              Nicolas Carion},
  title     = {{MDETR} - Modulated Detection for End-to-End Multi-Modal Understanding},
  journal   = {CoRR},
  volume    = {abs/2104.12763},
  year      = {2021},
  url       = {https://arxiv.org/abs/2104.12763},
  eprinttype = {arXiv},
  eprint    = {2104.12763},
  bibsource = {dblp computer science bibliography, https://dblp.org}
}

@article{DBLP:journals/corr/abs-1811-08342,
  author    = {Pravendra Singh and
               Manikandan R and
               Neeraj Matiyali and
               Vinay P. Namboodiri},
  title     = {Multi-layer Pruning Framework for Compressing Single Shot MultiBox
               Detector},
  journal   = {CoRR},
  volume    = {abs/1811.08342},
  year      = {2018},
  url       = {http://arxiv.org/abs/1811.08342},
  eprinttype = {arXiv},
  eprint    = {1811.08342},
  bibsource = {dblp computer science bibliography, https://dblp.org}
}

@article{doi:10.1177/1550147718764639,
author = {Lisang Liu and Fenqiang Liang and Jishi Zheng and Dongwei He and Jing Huang},
title ={Ship infrared image edge detection based on an improved adaptive Canny algorithm},
journal = {International Journal of Distributed Sensor Networks},
volume = {14},
number = {3},
pages = {1550147718764639},
year = {2018},
doi = {10.1177/1550147718764639},

URL = { 
        https://doi.org/10.1177/1550147718764639
    
},
eprint = { 
        https://doi.org/10.1177/1550147718764639
    
}
}

@article{DBLP:journals/corr/ZhuPIE17,
  author    = {Jun{-}Yan Zhu and
               Taesung Park and
               Phillip Isola and
               Alexei A. Efros},
  title     = {Unpaired Image-to-Image Translation using Cycle-Consistent Adversarial
               Networks},
  journal   = {CoRR},
  volume    = {abs/1703.10593},
  year      = {2017},
  url       = {http://arxiv.org/abs/1703.10593},
  eprinttype = {arXiv},
  eprint    = {1703.10593},
  bibsource = {dblp computer science bibliography, https://dblp.org}
}

@article{DBLP:journals/corr/IsolaZZE16,
  author    = {Phillip Isola and
               Jun{-}Yan Zhu and
               Tinghui Zhou and
               Alexei A. Efros},
  title     = {Image-to-Image Translation with Conditional Adversarial Networks},
  journal   = {CoRR},
  volume    = {abs/1611.07004},
  year      = {2016},
  url       = {http://arxiv.org/abs/1611.07004},
  eprinttype = {arXiv},
  eprint    = {1611.07004},
  bibsource = {dblp computer science bibliography, https://dblp.org}
}

@InProceedings{10.1007/978-3-642-13208-7_54,
author="Kursun, Olcay
and Alpaydin, Ethem",
editor="Rutkowski, Leszek
and Scherer, Rafa{\l}
and Tadeusiewicz, Ryszard
and Zadeh, Lotfi A.
and Zurada, Jacek M.",
title="Canonical Correlation Analysis for Multiview Semisupervised Feature Extraction",
booktitle="Artificial Intelligence and Soft Computing",
year="2010",
publisher="Springer Berlin Heidelberg",
address="Berlin, Heidelberg",
pages="430--436",
isbn="978-3-642-13208-7"
}

@article{DBLP:journals/corr/LimmerL16,
  author    = {Matthias Limmer and
               Hendrik P. A. Lensch},
  title     = {Infrared Colorization Using Deep Convolutional Neural Networks},
  journal   = {CoRR},
  volume    = {abs/1604.02245},
  year      = {2016},
  url       = {http://arxiv.org/abs/1604.02245},
  eprinttype = {arXiv},
  eprint    = {1604.02245},
  bibsource = {dblp computer science bibliography, https://dblp.org}
}

@inproceedings{inproceedings,
author = {Dou, Mingsong and Zhang, Chao and Hao, Pengwei and Li, Jun},
year = {2007},
month = {11},
pages = {722-732},
title = {Converting Thermal Infrared Face Images into Normal Gray-Level Images},
isbn = {978-3-540-76389-5},
doi = {10.1007/978-3-540-76390-1_71}
}

@misc{makeml,
  title        = "Ships Dataset",
  howpublished = "https://makeml.app/datasets/ships",
  journal      = "Make ML"
  }

@InProceedings{Ashraf_2021_CVPR,
    author    = {Ashraf, Muhammad Waseem and Sultani, Waqas and Shah, Mubarak},
    title     = {Dogfight: Detecting Drones From Drones Videos},
    booktitle = {Proceedings of the IEEE/CVF Conference on Computer Vision and Pattern Recognition (CVPR)},
    month     = {June},
    year      = {2021},
    pages     = {7067-7076}
}

@misc{rax2020,
  author       = "Robotic Automation Expert",
  title        = {"Introducing transfer learning as your next engine to drive future innovations"},
  howpublished = "https://medium.datadriveninvestor.com/introducing-transfer-learning-as-your-next-engine-to-drive-future-innovations-5e81a15bb567",
  month        = "Feb",
  year         = "2020",
  journal      = "Medium",
%   publisher    = "DataDrivenInvestor"
}

@article{article,
author = {Shoieb, Doaa and Youssef, Sherin and Aly, Walid},
year = {2016},
month = {12},
pages = {116-121},
title = {Computer-Aided Model for Skin Diagnosis Using Deep Learning},
volume = {4},
journal = {Journal of Image and Graphics},
doi = {10.18178/joig.4.2.122-129}
}

@INPROCEEDINGS{birdsAI,

  author={Bondi, Elizabeth and Jain, Raghav and Aggrawal, Palash and Anand, Saket and Hannaford, Robert and Kapoor, Ashish and Piavis, Jim and Shah, Shital and Joppa, Lucas and Dilkina, Bistra and Tambe, Milind},

  booktitle={2020 IEEE Winter Conference on Applications of Computer Vision (WACV)}, 

  title={BIRDSAI: A Dataset for Detection and Tracking in Aerial Thermal Infrared Videos}, 

  year={2020},

  volume={},

  number={},

  pages={1736-1745},

  doi={10.1109/WACV45572.2020.9093284}}

@inproceedings{HumanDetectionThermal,
author = {Ivašić-Kos, Marina and Krišto, Mate and Pobar, Miran},
year = {2019},
month = {04},
pages = {20-24},
title = {Human Detection in Thermal Imaging Using YOLO},
doi = {10.1145/3323933.3324076}
}

@inproceedings{ObjectDetectionThermal,
author = {Chaverot, Maxence and Carré, Maxime and Jourlin, Michel and Bensrhair, Abdelaziz and Grisel, R.},
year = {2021},
month = {01},
pages = {207-212},
title = {Object Detection on Thermal Images: Performance of YOLOv4 Trained on Small Datasets},
doi = {10.14428/esann/2021.ES2021-130}
}

@INPROCEEDINGS{ImageSegThermal,

  author={Chen, Yung-Yao and Chen, Wei-Sheng and Ni, Hui-Sheng},

  booktitle={2016 IEEE International Conference on Industrial Technology (ICIT)}, 

  title={Image segmentation in thermal images}, 

  year={2016},

  volume={},

  number={},

  pages={1507-1512},

  doi={10.1109/ICIT.2016.7474983}}

@unknown{ThermalSurvey,
author = {Kütük, Zülfiye and Algan, Görkem},
year = {2022},
month = {05},
pages = {},
title = {Semantic Segmentation for Thermal Images: A Comparative Survey},
doi = {10.48550/arXiv.2205.13278}
}

@article{WeaklyIntro,
author = {Zhou, Zhi-Hua},
year = {2017},
month = {08},
pages = {},
title = {A Brief Introduction to Weakly Supervised Learning},
volume = {5},
journal = {National Science Review},
doi = {10.1093/nsr/nwx106}
}

@article{MSCOCO,
  author    = {Tsung{-}Yi Lin and
               Michael Maire and
               Serge J. Belongie and
               Lubomir D. Bourdev and
               Ross B. Girshick and
               James Hays and
               Pietro Perona and
               Deva Ramanan and
               Piotr Doll{\'{a}}r and
               C. Lawrence Zitnick},
  title     = {Microsoft {COCO:} Common Objects in Context},
  journal   = {CoRR},
  volume    = {abs/1405.0312},
  year      = {2014},
  url       = {http://arxiv.org/abs/1405.0312},
  eprinttype = {arXiv},
  eprint    = {1405.0312},
  bibsource = {dblp computer science bibliography, https://dblp.org}
}

@misc{projectmaven,
author={Department of Defense Inspector General Office},
title={Evaluation of Contract Monitoring and Management for Project Maven},
address={4800 Mark Center Drive Alexandria, Virginia 223500-1500},
howpublished={DOD Report No. DODIG-2022-049},
year={2022},
url={https://media.defense.gov/2022/Jan/10/2002919460/-1/-1/1/DODIG-2022-049.REDACTED.PDF},}

\begin{table*}[ht]

\hskip-1cm
\begin{tabular}{|l|lll|ll|}
\multicolumn{1}{c}{}&\multicolumn{3}{c}{\large\textbf{Success}}&\multicolumn{2}{c}{\large\textbf{Failure}}\\
\hline
Ground Truth\label{tab:GT}&\includegraphics[width=\scales\linewidth,valign=m]{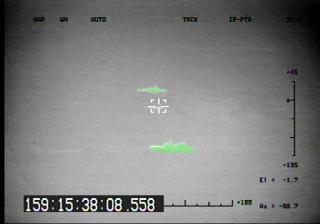} & \includegraphics[width=\scales\linewidth,valign=m]{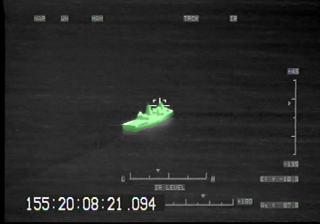} &\includegraphics[width=\scales\linewidth,valign=m]{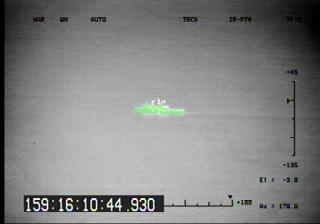}  & \includegraphics[width=\scales\linewidth,valign=m]{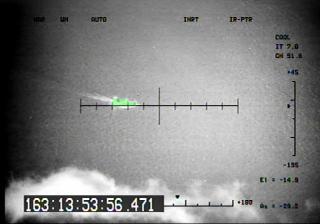}&\includegraphics[width=\scales\linewidth,valign=m]{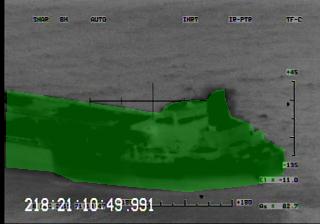}\\
N=10 &\includegraphics[width=\scales\linewidth,valign=m]{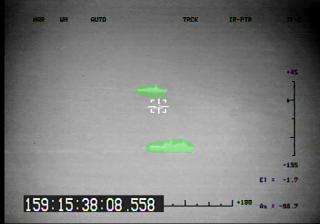} & \includegraphics[width=\scales\linewidth,valign=m]{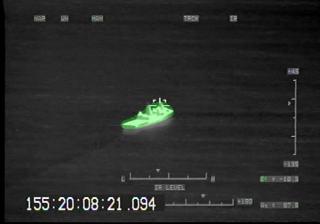} &\includegraphics[width=\scales\linewidth,valign=m]{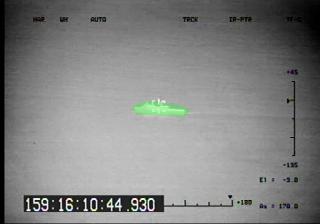}& \includegraphics[width=\scales\linewidth,valign=m]{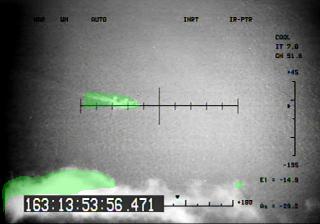}&\includegraphics[width=\scales\linewidth,valign=m]{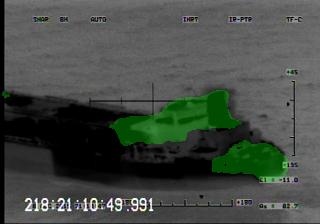}
\\
N=32 &\includegraphics[width=\scales\linewidth,valign=m]{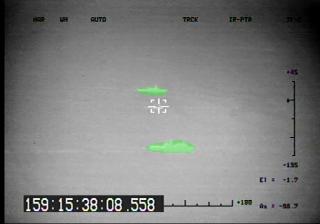} & \includegraphics[width=\scales\linewidth,valign=m]{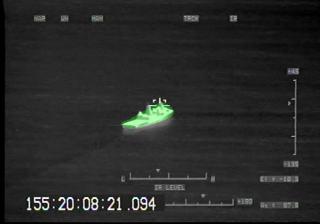} &\includegraphics[width=\scales\linewidth,valign=m]{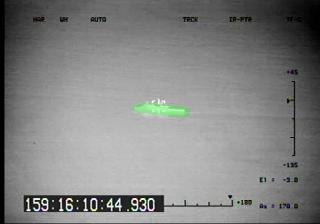}& \includegraphics[width=\scales\linewidth,valign=m]{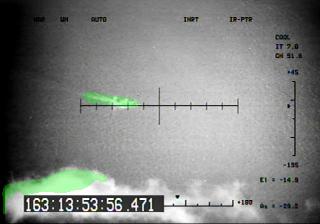}&\includegraphics[width=\scales\linewidth,valign=m]{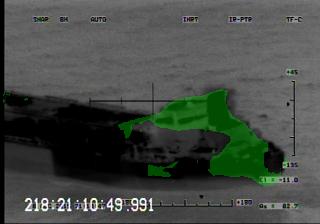}\\
\hline
\end{tabular}
\caption{\small Success and failure cases of segmentation mask predictions, using the two different annotation schemes.}
\label{tab:images}

\end{table*}
\begin{table*}[ht]
    \centering
    \begin{tabular}{|c|c|c|c|c|}
         \hline
         \textbf{Supervision}&\textbf{Augmentations}&\textbf{Precision}&\textbf{Recall}&\textbf{Jaccard Score}\\
         \hline
         Squiggle (n=32)&None&.925&.948&.612\\
         \hline
         Squiggle (n=32)&Grayscale and Inversion&.979&.978&.756\\
         \hline
         Point (n=10)&Grayscale and Inversion&.976&.976&.692\\
        \hline

    \end{tabular}
    
    \caption{\small Calculated Precision, Recall and Jaccard scores of the inference image mask predictions made by a network pre-trained on the Airbus dataset, scored against a lmited set of test images with full segmentation maps.}\vspace{10pt}
    \label{tab:pretrained}
    \vspace{100pt}
    \begin{tabular}{|c|c|c|c|c|}
         \hline
         \textbf{Supervision}&\textbf{Augmentations}&\textbf{Precision}&\textbf{Recall}&\textbf{Jaccard Score}\\
         \hline
         Squiggle (n=32)&None&.891&.922&.601\\
         \hline
         Squiggle (n=32)&Grayscale and Inversion&.972&.973&.726\\
         \hline
         Point (n=10)&Grayscale and Inversion&.961&.962&.687\\
        \hline
    \end{tabular}
    \caption{\small Calculated Precision, Recall and Jaccard scores of the inference image mask predictions made by a non-pretrained network, scored against a lmited set of test images with full segmentation maps.}
    \label{tab:not-pretrained}
\end{table*}


\newpage

\end{document}